\documentclass[a4paper,conference]{IEEEtran}
\ifCLASSINFOpdf
\else
\fi
\usepackage{cite}
\usepackage{amsmath,amssymb,amsfonts}
\usepackage{graphicx}
\usepackage{textcomp}
\usepackage{xcolor}
\usepackage{ dsfont }
\usepackage{times}
\usepackage{epsfig}
\usepackage{subcaption}
\usepackage{graphicx}
\usepackage{amsmath}
\usepackage{amssymb}
\usepackage{algorithm}
\usepackage{booktabs}
\usepackage[noend]{algpseudocode}
\usepackage{amsmath}
\DeclareMathOperator*{\argmax}{arg\,max}
\DeclareMathOperator*{\argmin}{arg\,min}

\makeatletter
\def\endthebibliography{%
  \def\@noitemerr{\@latex@warning{Empty `thebibliography' environment}}%
  \endlist
}
\makeatother

\begin{document}
%
\title{Prb-GAN: A Probabilistic Framework for GAN Modelling}

\author{\IEEEauthorblockN{ Blessen George}
\IEEEauthorblockA{ 
\textit{Indian Institute of Technology Kanpur}\\
Kanpur, India \\
    george@iitk.ac.in}
\and
\IEEEauthorblockN{ Vinod K. Kurmi}
\IEEEauthorblockA{ 
\textit{Indian Institute of Technology Kanpur}\\
Kanpur, India \\
    vinodkk@iitk.ac.in}
\and
\IEEEauthorblockN{ Vinay P. Namboodiri}
\IEEEauthorblockA{ 
\textit{Indian Institute of Technology Kanpur}\\
Kanpur, India \\
vinaypn@iitk.ac.in}

}

\maketitle

\begin{abstract}

Generative adversarial networks (GANs) are very popular to generate realistic images, but they often suffer from the training instability issues and the phenomenon of mode loss. In order to attain greater diversity in GAN synthesized data, it is critical to solving the problem of mode loss. Our work explores probabilistic approaches to GAN modelling that could allow us to tackle these issues. We present Prb-GANs, a new variation that uses dropout to create a distribution over the network parameters with the posterior learnt using variational inference. We describe theoretically and validate experimentally using simple and complex datasets the benefits of such an approach. We look into further improvements using the concept of uncertainty measures. Through a set of further modifications to the loss functions for each network of the GAN, we are able to get results that show the improvement of GAN performance. Our methods are extremely simple and require very little modification to existing GAN architecture.
\end{abstract}


\section{Introduction}

Generative modelling is a branch of machine learning research that aims at learning the underlying probability distribution of a data source given a few samples from it. Generative Adversarial Networks (GANs)~\cite{gan} are one such class of generative models that, especially in the context of computer vision, are capable of producing highly realistic data. The model is based on a game theoretic training process that uses two neural networks termed the discriminator and the generator. This scheme, however, is often susceptible to training instability along with other pathological problems like vanishing gradients, mode dropping and hyperparameter initialization sensitivity. 

Since GANs are quite a popular generative model, it is critical to solve the issues of mode loss and training instability and thus forms an active area of GAN research. Multiple different approaches to these problems have been tried in past works. Multiple networks based methods such as \cite{gman},~\cite{ghosh2018multi}, ~\cite{dropoutgan} are all very effective to tackle these issues. The underlying idea behind all these methods is that, from multi networks, the generator can capture the data distributions efficiently. 
These models are trained by  a deterministic loss function obtained from the discriminator. The probabilistic loss-based approach could also one of the solutions to tackle the problem of mode collapse. The fact is that the probabilistic models incorporate the variance estimation; thus, they can provide better gradients to update the generator model. The main issue with the probabilistic methods is that they are intractable. But Gal et al. ~\cite{dropoutbayesianapprox} have proved a method by applying the dropout to obtain the Bayesian formulation of the network. The dropoutGAN~\cite{dropoutgan} is one of such type of variations of the GAN. But this work mainly involves using multiple discriminators and a dropout branch to choose a subset of the discriminators to train the generator against in each iteration. Our work, on the other hand, involves applying dropout to the parameters of the discriminator network actually to simulate a probabilistic discriminator. 
 The spirit of our work is very different from the dropoutGAN~\cite{dropoutgan}. 
In all of these works, the question remains as to exactly how many multiple networks needed, and this generally must be estimated via trial and error. In contrast, we use just a single discriminator and a single generator network; thus, our memory footprint is much smaller. 

Another important aspect of the Bayesian formulation of GAN is to incorporate the uncertainty estimation. In the proposed framework, we also used the uncertainty estimation for the measurement of the GAN loss. There are two variant of this uncertainty based measures are \textit{Weighted Discriminator Score} and \textit{Discriminator Score Set Variance} used to obtain the new GAN losses. In this paper, we work with predictive uncertainty~\cite{uncertainty2} of the deep learning model using the dropout.

We have drawn our inspiration from BayesianGAN~\cite{bayesiangan}  and concur that thinking in a probabilistic manner would yield fruits in improving GAN performance. While the direction of our work runs similar to~\cite{bayesiangan}, our implementation is extremely simple and requires minimal extra machinery. 
\vspace{1em}

Our contributions are as follows.
\begin{itemize}
    \item We propose a variation call it Prb-GAN (Probabilistic GANs) which uses the popular technique of Dropout~\cite{srivastava2014dropout} to convert standard GANs into probabilistic GANs. Inference of the parameter distribution is done using variational inference.
    \item We also present the theoretical backing for our work and show empirically why such a formulation can help solve the mode loss issue.
    \item Additionally, we present new loss functions that are based on uncertainty measure estimates that further improve the GAN performance metric. 
\end{itemize}

\vspace{2em}

\section{Related Work}

In deep learning literature, many GAN architectures and models have been proposed. One line of work is to change the objective itself entirely as proposed by works such as Least squares GANs (LSGANs)~\cite{mao2017least} and Wasserstein GANs~\cite{maxslicedwgan,slicedwgan_ishan,slicedwgan_luc,gulrajani2017improved}. The assertion is that the standard objective function minimizes the JS-divergence, which is unstable, particularly when the support of both distributions have no intersection. By minimizing the more smoother Wasserstein distance, one is able to significantly check the instability. LSGANs are similar in approach in that it minimizes the Pearson divergence. MADGANs~\cite{madgan} was another recent approach wherein the authors advocate the use of multiple generators, each tasked with capturing a separate mode of the data. Bayesian GANs are the only GAN variations that approached the problem in a Bayesian manner~\cite{bayesiangan}. The key idea here was to maintain a distribution over the parameters of the networks and to infer the right posterior distribution given the observed samples from the dataset. Another probabilistic version of GAN is P-GAN~\cite{eghbal2017probabilistic}, where instead of drawing a boundary between real and fake samples, discriminator of P-GAN fits a Gaussian on the embeddings of the projected real images. The f-GAN~\cite{nowozin2016f} summarizes that GANs can be trained by using an f-divergence.  LS-GAN~\cite{qi2019loss} is introduced to train the generator to produce
realistic samples by minimizing the designated margins
between real and generated samples. MRGAN~\cite{che2017mode} proposes a metric regularization to penalize missing modes. SN-GAN~\cite{miyato2018spectral} proposes the use of weight normalization to stabilize the training of the discriminator.

Apart from the different loss-variant, there are many other architectures based GAN models are proposed. The BGAN~\cite{hjelm2018boundary}, PROGAN~\cite{gao2019progan}, BigGAN~\cite{donahue2019large}, StarGAN~\cite{choi2018stargan} and many others are popular to generated the realistic images.
Recently there are many other variants and applications of GANs~\cite{Agustsson_2019_ICCV,Yang_2019_ICCV,Gong_2019_ICCV,Shaham_2019_ICCV,Kundu_2019_ICCV,Huang_2019_CVPR,Mao_2019_CVPR,Eghbal-zadeh_2019_CVPR,Jenni_2019_CVPR} in different areas are presented in the literature. Recently some probabilistic version of GANs are present in ProbGAN~\cite{probgan_iclr} where distributions of the generator and the discriminator are leaned. In IDDA~\cite{kurmi2019looking} a class based discriminator for learning the source distribution. In the PCGAN~\cite{han2020robust}, authors used a conditional generative framework along with  Pairwise Comparisons using the uncertainty estimation.

In addition to the GAN literature, the probabilistic and Bayesian deep learning-based models are also popular. In~\cite{dropoutbayesianapprox} Gal et al. propose that through the use of dropout~\cite{srivastava2014dropout} in neural networks, one could create a Bayesian neural network. By requiring a prior distribution over the weights of the neural networks and training using dropout in the layers of the networks, we could infer the distribution of the weights of the neural network. The technique offers a clean way of creating probabilistic neural networks. We use this idea and the assertion that GANs could indeed benefit from maintaining a distribution over the network parameters to try and solve the mode collapse and stability issues prevalent in GANs. The uncertainty estimation aleatoric and epistemic for the deep learning models are presented in~\cite{uncertainty2, kurmi2019attending}. Other uncertainty based methods are discussed in~\cite{NIPS2017_7219,hendrycks2019augmix,kurmi2019curriculum}.




    \section{Probabilistic GANs}
    \subsection{Standard GANs}
        Recall that the original GAN formulation features a generator and a discriminator with a single parameter setting $\theta_g \text{ and } \theta_d$ respectively rather than a parameter distribution. The optimal parameter point estimate is found by solving  
    \begin{gather}
		\theta_g^* = \max_{\theta_g} \enskip\log D(G(z)) \\
		\theta_d^* = \min_{\theta_d} \enskip[\log D(x) + \log [1-D(G(z))]]         
    \end{gather}
    Here $z$ is the latent noise vector.
        We show in the next sections the simple changes required to convert the standard GAN into a probabilistic GAN. 
    
	\subsection{Probabilistic Generators and Discriminators}
    \textbf{Probabilistic Generators:}
    Assume a gaussian prior distribution over the weights of generator $\theta_g$, where $n_{\theta_g}$ represents number of generator parameters, 
	\begin{gather}
		p(\theta_g) \sim \mathcal{N}(0, I_{n_{\theta_g}})
	\end{gather}
	Let the probability that the data point produced by the generator $G(z; \theta_g)$ being marked real by the discriminator be given by 
	\begin{gather}
		p(y_g = 1 | \theta_g, z) = D(G(z; \theta_g);\theta_d)
	\end{gather}
	Here $\theta_d$ are the parameters of discriminator network.
	Then we seek to find the posterior distribution of the weights $p(\theta_g | y_g = 1, z)$ assuming that the generated point was marked real. Direct computation is infeasible and we infer using variational inference. 
	
	We setup our variational distribution for the generator weights using Bernoulli dropout as 
	\begin{gather}
	q(\theta_g; W_g) = \{\theta_g^i\} = \{W_g^i . \text{diag}([b_{i, j}]_{j=1}^{K_i})\}  \\
	b_{i,j} \sim \text{Bernouilli}(p_i) ; i \in [L], j \in [K_i]
	\end{gather}
	 
	Here $W_g$ serves as the variational parameter, $K_i$ is the number of neurons in the $i^{th}$ layer, $L$ is the number of layers and $p$ is the dropout probability. We seek out the optimal variational parameters for the generator by maximizing the evidence lower bound (ELBO) function associated with the above. Thus 
	\begin{gather}
	\begin{split}
		W_g^*  = & \argmax_{W_g} \enskip E_{q(\theta_g; W_g)} \enskip[\log p(y_g = 1 | \theta_g, z)] \\
		&\quad - KL (q(\theta_g; W_g) || p(\theta_g))
	\end{split}
	\end{gather}

	The first term is an expectation and can be approximated using Monte-Carlo(MC)~\cite{hastings1970monte} integration for some arbitrary MC iteration, $M_g$, and the second term corresponds to the $L_2$ norm regularization term which forces the parameters to stay close to 0. Thus the loss function we would like to minimize becomes 
	\begin{gather}
		W_g^* = \argmin_{W_g} \left\lVert W_g\right\rVert^2_2 - \dfrac{1}{M_g}\sum_{m_g=1}^{M_g} \log D(G(z; \theta_g^{(m_g)}))
	\end{gather}
\textbf{Probabilistic Discriminators:}
	The setup for probabilistic discriminators are nearly the same. We place a gaussian prior over the discriminator parameters as well. The likelihood term however is slightly different. The probability that the discriminator marks the real generated points as real and fake respectively appropriately is: 
	\begin{gather}
		p(y_d = [1,0] | \theta_d, x, z) = D(x; \theta_d).(1 - D(G(z); \theta_d))
	\end{gather}
	Then, the posterior distribution of the discriminator weights is sought out using a similar bernouilli dropout based variational distribution $q(\theta_d; W_d)$. The loss function that follows is given by 
	\begin{gather}
	\begin{split}
		W_d^* = \argmin_{W_d} \left\lVert W_d\right\rVert^2-2 - & \dfrac{1}{M_d}\sum_{m_d=1}^{M_d} \enskip [\log D(x; \theta_d^{(m_d)}) \\
		& + \log(1-D(G(z); \theta_d^{(m_d)}))]
	\end{split}
	\end{gather}
	
	$M_d$ is the number of sampled discriminator.

	\begin{figure}[htb]
		\centering
		\includegraphics[width=0.45\textwidth]{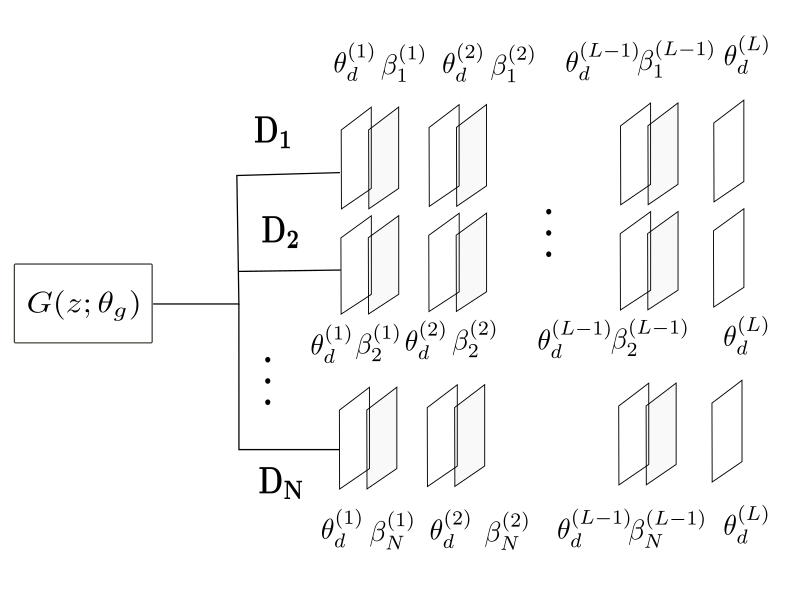}
		\caption{
		    During a training step of the probabilistic discriminator, $N$ discriminator instances and a single random generator are sampled. The losses are averaged and back-propogated to update the discriminator variational parameter.
		}
		\label{fig:dis}
	\end{figure}

	\begin{figure}[htb]
		\centering
		\includegraphics[width=0.45\textwidth]{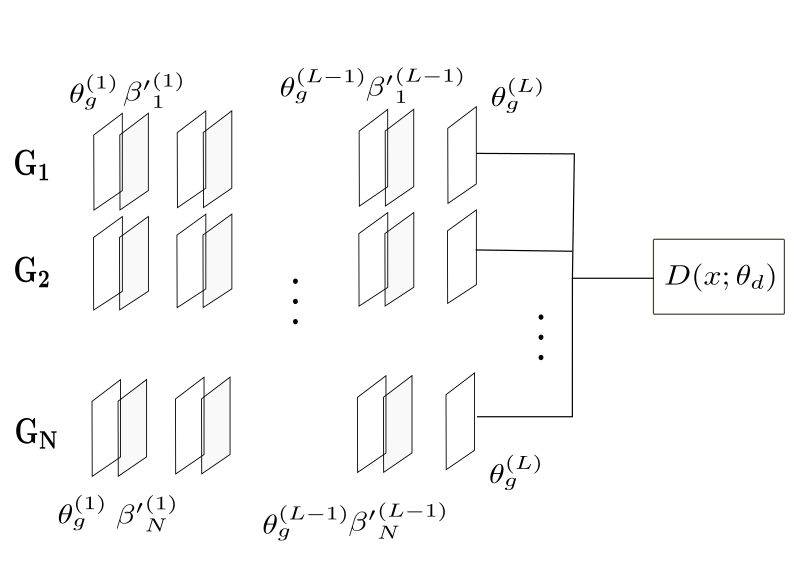}
		\caption{$N$ generator instances and a single random discriminator are sampled and the losses backpropogated during one generator training step.}
		\label{fig:gen}
	\end{figure}

\textbf{Training scheme}: In practice, the updated loss function for the generator is equivalent to actually sampling the probabilistic generator $M_g$ many times and back-propagating the averaged loss for a single generator training step and similarly sampling the discriminator $M_d$ many times and averaging the loss during a single discriminator training step. For each discriminator training step, a single random generator is sampled and vice versa for a single generator training step. Generally, we set the number of MC samples for generator and discriminator to be equal $M_d = M_g = N$. Figure~\ref{fig:dis} and ~\ref{fig:gen} describes how the training steps would proceed each case. For a generator training step, while the variational parameters $\theta_g^{(l)}$ are the same for all generators, unique sampling of dropout layers ${\beta^\prime}_n^{(l)}$ create a unique generator. The case holds similarly for the discriminator training step. The algorithm is presented in  ~\ref{alg:algorithm1}. We term our formulation of GAN as Prb-GAN. \\


	\begin{algorithm}
	\caption{Probabilistic GAN}\label{alg:dropoutgan}
	\begin{algorithmic}[1]
	\State $\text{Initialize variational parameters } W_d, W_g \text{ randomly}$
	\State $\text{Let $B$: Batch Size, $N$: MC sample iterations}$
	\While{not converged }\\
	\State $\{z^i\} \sim p(z); \: {i \in [B]}$ \Comment{Sample noise variables}
	\State $\{x^i\} \sim X; \: {i \in [B]}$ \Comment{Sample real data samples}
	\\
	\State ${\theta_g}_{*} \sim q({W_g}; p)$ \Comment{Choose random generator network parameter}

	\For {i=1 to N}
	\State ${\theta_d}^{(i)} \sim q({W_d}; p)$ \Comment{Sample a discriminator}
	\State $f_d \gets \sum_{b=1}^B [\log D(x^i; {\theta_d}^{(i)}) + \log(1-D(G(z^i; {\theta_g}_{*}); {\theta_d}^{(i)}) ]$ \Comment{Loss}
	\State $\alpha_i \gets \nabla_{{\theta_d}^{(i)}}f_d$ \Comment{Gradient}
	\EndFor
	\State $W_d \gets W_d + \dfrac{\lambda}{BN} \sum_{i=1}^N \alpha_i$ \Comment{Update using average gradient}
	\\
	\State ${\theta_d}_{*} \sim q({W_d}; p)$ \Comment{Choose random discriminator network parameter}

	\For {i=1 to N}
	\State ${\theta_g}^{(i)} \sim q({W_g}; p)$ \Comment{Sample a generator}
	\State $f_g \gets \sum_{b=1}^B [\log(D(G(z^i; {\theta_g}^{(i)}); {\theta_d}_{*}) ]$ \Comment{Loss}
	\State $\beta_i \gets \nabla_{{\theta_g}^{(i)}}f_g$ \Comment{Gradient}
	\EndFor
	\State $W_g \gets W_g + \dfrac{\lambda}{BN} \sum_{i=1}^N \beta_i$ \Comment{Update using average gradient}
	\\

	\EndWhile\label{dropoutgan}

	\State \textbf{return} $W_d, W_g$

	\end{algorithmic}
	\label{alg:algorithm1}
	\end{algorithm}
    \vspace{-1.0em}
    \section{Uncertainty Measures for GANs}
    \label{sec:uncert}
In this section, we make a further improvement to our approach through the use of uncertainty measures for GANs. Kendall et al. ~\cite{uncertainty,uncertainty2} discussed how neural networks could be equipped with prediction uncertainty estimates in addition to their actual predictions on regression and classification tasks. We work with two types of uncertainty measures - Aleatoric uncertainty, which is the uncertainty that arises due to the data itself and Epistemic uncertainty, which arises when the model has not been trained well enough on the data. Aleatoric uncertainty could be occlusion in the data, multiclass instances in a single image, noise due to the data capturing equipment itself, or any other problem that causes input data to not conform to the expected standard. In such cases, even if our model is trained to the best of its ability, it may still not be certain about its predictions. Essentially, this is the uncertainty that cannot be 'trained' away with lots of data. Epistemic uncertainty measures the uncertainty about prediction based on how well the model was trained. For example, suppose the data is perfectly free from noise, assume a prediction model that has been trained only on a very small portion of the observed data. Due to the lack of training, the model cannot be certain about its prediction, and thus the predictive posterior may have a widespread. As more and more data is seen, the variance in the predictive posterior decreases and concentrated towards its mean. Thus we say that as the model is trained, it is increasingly free of epistemic uncertainty.

   Inspired by these ideas, we asked whether GANs could benefit from a similar line of thought? Essentially, we ask \textit{Could we equip discriminators with the ability to be certain or uncertain about the predictions? And could we use the variance of the MC sampled discriminator scores to improve the generator's gradient? }
    
    \subsection{Weighted discriminator scores}

\begin{figure}[htb]
		\centering
		\includegraphics[scale=0.28]{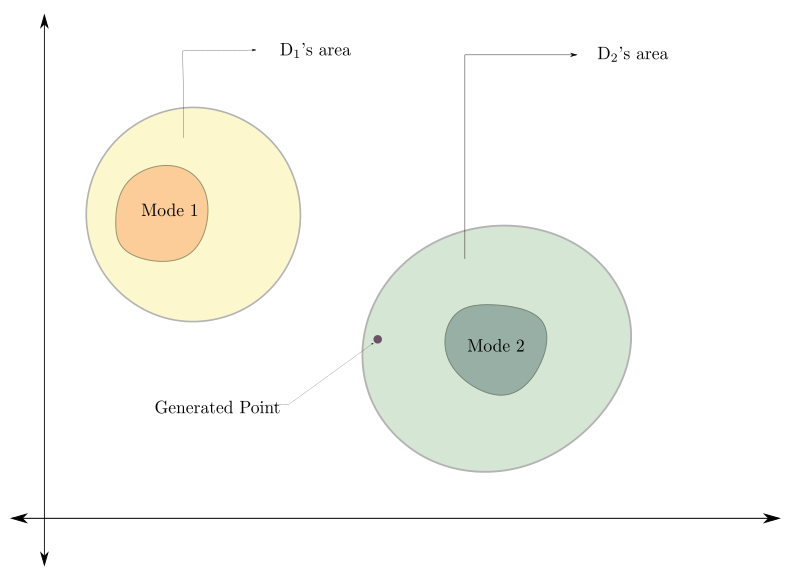}
		\vspace{-0.5em}
	    \caption{Consider two modes captured by two separate discriminators. In the case that the input sample lies close to the second mode, discriminator 2 could choose to be more certain about its prediction while discriminator 1 could choose to be less certain.}
	    \label{fig:aleatoric_uncertainty}
	    \vspace{-1em}
	\end{figure}

    Inspired by the idea of determining aleatoric uncertainty for neural nets, we allow each dropout sampled discriminator to decide on how certain or uncertain it wants to be about its prediction on its input data-point. Every randomly sampled discriminator corresponds to instantiating a decision boundary for separating real and fake data points. When input data-points are closer to the captured mode of one of these sampled discriminators, if that discriminator makes a prediction with high certainty, its score is weighted higher. This encourages discriminators to capture diverse modes of the data by allowing it to be sure of what it knows and unsure of what it does not know. Figure~\ref{fig:aleatoric_uncertainty} illustrates that since the input datapoint is closer to discriminator 2's captured modes, we would like the discriminator 2 to be more sure about its prediction as compared to discriminator 1. 
	
	Following this, we create a new loss function that essentially weighs the prediction scores of the discriminators based on their uncertainty scores. Uncertainty scores are estimated by adding another branch from the penultimate layer of the discriminator.  Greater the uncertainty measure, the closer the predicted logit score of that discriminator is to 0 (correspondingly the actual prediction being close to 0.5).  
	Consider the logits predicted by a discriminator $D(x; \theta_d)$. Suppose we allow the discriminator to specify how certain it is about its prediction - uncertainty represented as $u(x)$, then the modification is  
	\begin{gather}
		D(x; \ \theta_d) \rightarrow D^\prime(x; \ \theta_d) := \dfrac{D(x; \ \theta_d)}{u(x; \ \theta_d) + b_1}
	\end{gather}
	Here $b_1$ is a hyperparameter that stands for a bias term which in we set a small number in our experiments. This is to prevent division by zero in case the uncertainty predicted is 0. 
	The effect of distorting the predicted logits as above is to increase or decrease the prediction magnitude thus making the logit prediction closer to 0 in case $u(x; \theta_d) > 1$ or further away from 0 otherwise. 

	To prevent the discriminator from assigning high uncertainty to all the points we add the uncertainty estimate itself as an additional penalty term. Thus the loss function for the discriminator factoring in this input data uncertainty is changed Eq.~\ref{eq:old} to Eq.~\ref{eq:new}, $\text{BCE}$ is the binary cross-entropy loss.  
	\begin{gather}
	\begin{split}
		\mathcal{L}_{\text{old dis loss}} = \dfrac{1}{N} \text{BCE}(\sigma (D(x; \ \theta_d)))
	\end{split}
	\label{eq:old}
	\end{gather}
	\begin{gather}
	\begin{split}
			\mathcal{L}_{\text{new dis loss}} = \dfrac{1}{N} [\text{BCE } & \left(\sigma \left(D^\prime(x; \ \theta_d) \right)\right) \\ 
		& \ + \ u(x; \ \theta_d)]
	\end{split}
		\label{eq:new}
	\end{gather}
$\text{BCE}$ is the binary cross-entropy loss.
	\subsection{Discriminator score set variance}

	Our second modification is based on the idea that we could measure the variance of the set of discriminator scores and use this information to improve gradients for the generator. Essentially, if the discriminator scores exhibit high variance, then the sampled discriminators must disagree with each other's scores and some discriminators must have assigned high logit scores while others must have scored the point with low logit scores. This would happen presumably when the data point is closer to modes captured by some of the discriminators while lying outside the high probability regions of the remaining discriminators which would indicate that the discriminators have captured diverse modes. When the generator generates such data points we'd like to reward it based on the magnitude of the score set variance. 

	Consider a concrete example wherein two separate cases, the discriminators assign logit scores to the same generated point as $[7, -1, 7, -1]$ and $[3, 3, 3, 3]$.  Now the mean score in both cases is equal to 3; however, the first case though, exhibits high variance among the scores while the second set does not. We argue that the first case might be more desirable due to the fact that one-half of the discriminators recognize the point as being close to their captured modes while the other half marks them as unsure or outside their captured modes thus again incentivizing diverse mode capture.  

To implement this idea, we introduce an additional term in the loss function of the generators that rewards generated points that have high variance in the discriminator logit score set. The updated loss function for the generator is given as
\vspace{-0.5em}
	\begin{gather}
		\begin{split}
		\mathcal{L}_{\text{new gen loss}} = \dfrac{1}{N}\sum_{n=1}^{N}\text{BCE}(D_n^\prime(G(z))) \\ -  \lambda \dfrac{var\{D_n\prime(G(z))\}_{n=1}^{N}}{\mu^2\{D_n\prime(G(z))\}_{n=1}^{N} + b_2}
		\end{split}
	\end{gather}

	  In the above $D_n$ represents the $n^{th}$ sampled discriminator. The term features $\lambda$ - a hyperparameter, $b_2$ - a fixed bias term to prevent division by 0, and the mean and variance of the set of discriminator scores. If the mean of the prediction is closer to zero, then the term has higher weightage. This is done to emphasize that only when the mean of the discriminator scores is near 0, is the term relevant. Greater the variance of the discriminator scores, the greater the reward.


\section{Results}

	We conduct two different sets of studies to measure the performance of Prb-GAN against other standard GANs. The first is a set of initial experiments we conducted to validate our ideas. The second is a larger study comparing the current best GAN evaluation metric - the Frechet Inception Distance (FID) score on more standard datasets - CIFAR 10 and Celeb-A datasets. Finally, we show the results for uncertainty measure equipped probabilistic GANs.

	\subsection{Small scale experiments}
Initially, we evaluated our model on a set of two simple datasets, a mixture of 1-dimensional Gaussians and the MNIST dataset. We show that on both the datasets, whereas the standard GAN exhibits mode loss behavior, the probabilistic formulation, as described above, is able to significantly prevent mode loss. We use a simple, fully connected 4 layer neural network with a net input size of $z_d=100$ where $z_d$ is the dimensionality of the random latent code space and hidden layer size of 600 neurons with the leaky Relu activation function. Xavier's initialization of weights is used. We set the probability of dropout $p$= 0.4 and $N$ = 20. 

 The 1D Gaussian dataset is comprised of data-points in $R$ sampled from a mixture of 5 Gaussians. The means of the Gaussians are located at $[10, 20, 60, 80, 110]$ with standard deviations as $[3, 3, 2, 2, 1]$. The histogram representing the real dataset is shown in red, while the generated data histogram is shown in green. This is the same dataset used in~\cite{madgan}. The dataset consists of 600,000 data-points sampled from the mixture distribution. An ideal generative model would be able to recreate the dataset perfectly, and the histogram associated would resemble the true distribution.
 The vanilla GAN and Prb-GAN are trained for 5 epochs over the entire dataset.

\begin{figure} 
		\centering
		\vspace{-1em}
	    \begin{subfigure}[!]{0.45\textwidth}
	    	\centering
	        \includegraphics[width=\linewidth]{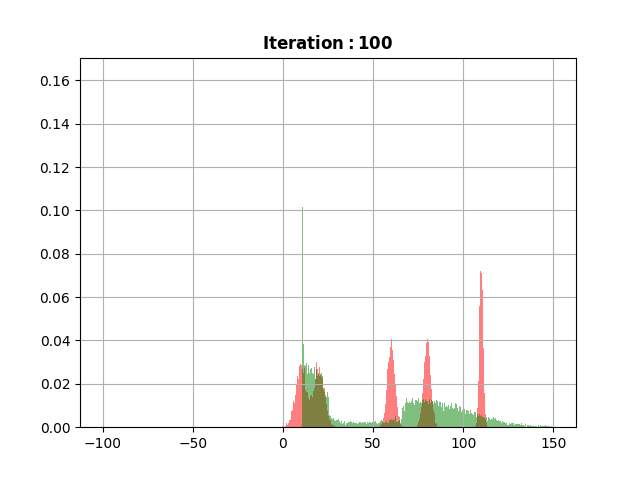}
	         \vspace{-2.8em}
	        \caption{Vanilla GAN}
	        \label{fig:vanilla_gaussian}
	    \end{subfigure} \\~\\
	    \begin{subfigure}[!]{0.45\textwidth}
	    	\centering
	        \includegraphics[width=\linewidth]{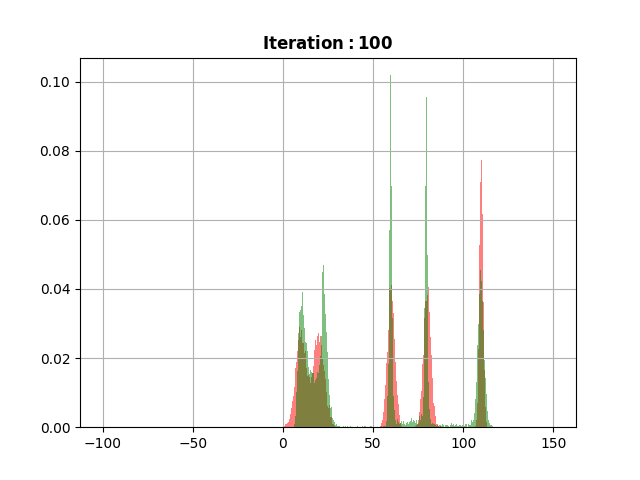}
	         \vspace{-2.8em}
	        \caption{Prb-GAN}
	        \label{fig:probabilistic_gaussian}
	    \end{subfigure}%
	    \caption{Histogram plots of the generated data (in green)  and the real data (in red). At 100 iterations, Prb-GAN is able to capture all the modes while Vanilla GAN is not. }
	    \vspace{-2em}
	    \label{fig:1dgaussian}
	\end{figure}

 Figure~\ref{fig:vanilla_gaussian} and~\ref{fig:probabilistic_gaussian} shows the histogram plots of the data generated by both the models. It is observed that while the vanilla GAN seems to capture only one mode well, the dropout GAN is able to capture almost all the modes. In fact, by iteration 50, we observed that Prb-GAN was able to capture four of the most important modes, while vanilla GAN was only able to capture one mode. Also, we noted that the values for dropout probability and $N$ (number of MC sample iterations) used can change the results. In general, the greater the dropout probability used, the greater the $N$ value that would have to be in order to produce good samples. 


     \begin{figure}
        \centering
	    \begin{subfigure}[!]{0.20\textwidth}
	        \includegraphics[width=\linewidth]{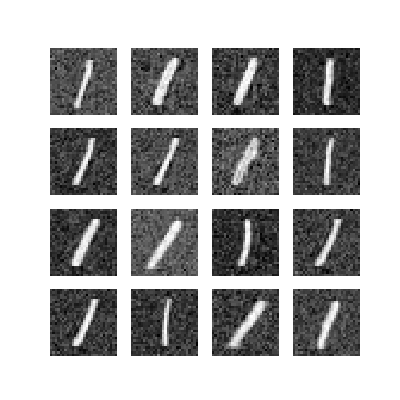}
	        \caption{Vanilla GAN}
	        \label{fig:vanilla_cifar10}
	    \end{subfigure}
	    \hspace{0.5cm}
	    \begin{subfigure}[!]{0.20\textwidth}
	        \includegraphics[width=\linewidth]{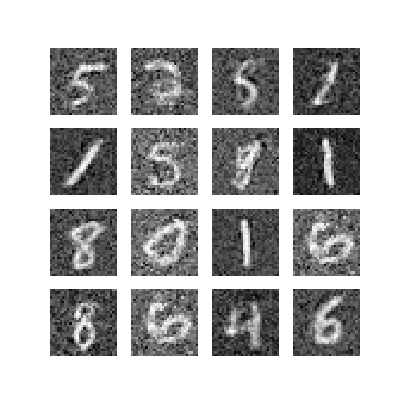}
	        \caption{Prb-GAN}
	        \label{fig:vanilla_cifar10}
	    \end{subfigure}
	    \caption{Generated images at 100 iterations. We observe that while Vanilla GAN has mode collapsed to the digit 1, Prb-GAN is able to produce diverse MNIST digits thus showing itself robust to mode collapse. }
	     \vspace{-1em}
	     \label{fig:mnist}
    \end{figure}
 
  \begin{figure}
        \centering
	    \begin{subfigure}[!]{0.20\textwidth}
	        \includegraphics[width=\linewidth]{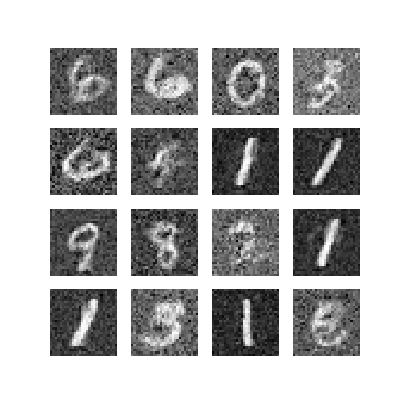}
	        \caption{Prb-GAN instance 1}
	        \label{fig:vanilla_cifar10}
	    \end{subfigure}
	    \hspace{0.5cm}
	    \begin{subfigure}[!]{0.20\textwidth}
	        \includegraphics[width=\linewidth]{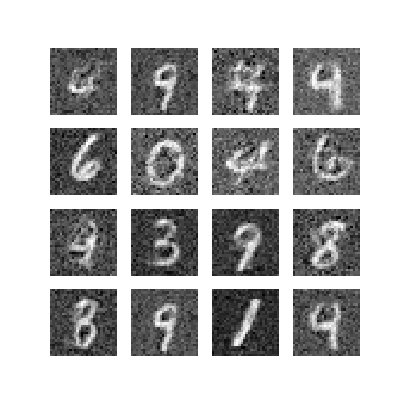}
	        \caption{Prb-GAN instance 2}
	        \label{fig:vanilla_cifar10}
	    \end{subfigure}
	    \caption{Using the same latent code for the generator with the same variational parameter, but sampled separately through dropout, we are able to produce diverse results. Prb-GAN instance 1 produces more 1's and Prb-GAN instance 2 produces more 9's as the digit output. }
	          \vspace{-1em}
      \label{fig:diff_outputs}
    \end{figure}

Following this experiment, we tested our model on real images, specifically the MNIST dataset. We use the same GANs but make a change to the input layer size to match the input latent code size, which we set to be 200. We run the models over the entire dataset for 100 epochs and plot the images generated by each model at the end of the training. Figure~\ref{fig:mnist}(a) and~\ref{fig:mnist}(b) shows how while Vanilla GAN has collapsed the entire latent code mapping to the digit 1, Prb-GAN is able to produce diverse images. We note that the visual quality of the Prb-GAN may appear slightly less appealing, but what we gain in return is larger diversity. Through the use of Progressive GANs~\cite{proggans}, one could conceive of using a Prb-GAN backend to allow the network to produce diverse images while having a standard Progressive GAN front end to make the produced images more crisp.  

	    

  Further, in Figure~\ref{fig:diff_outputs}, we see that using the exact same random latent code, Prb-GAN in separate Bernoulli dropout sampling instances is able to produce very different images from each other. Thus separate instances of the generators actually capture different sets of modes. This observation shows that through the use of a single GAN model and the dropout technique, we're able to achieve performance similar to peers like MADGAN while having the advantage of not having to know in advance the number of modes that could be present in the data. 

	\subsection{Standard GAN metric evaluation study}

	 To test our GAN performance more rigorously, we turn towards the work done by \cite{comparegan}. Their work compares all the standard GANs on various datasets on a variety of hyperparameters and measures them using the FID score. As opposed to the earlier used Inception score~\cite{inception}, the FID score~\cite{fid} is the current best GAN evaluation metric since it is more robust to noise and sensitive to mode loss compared to Inception score. The surprising conclusion reached is that it is possible that all the tested GANs perform nearly equal. We demonstrate that while this might be the case, the use of our probabilistic approach in GANs almost always improves performance.

  Our architectural setup is as used in CompareGAN~\cite{WinNT}. We compare three of the best variations of GANs - Non-Saturating GANs (NSGANs, the original GAN)~\cite{gan} and  Least-squares GANs (LSGANs)~\cite{lsgan}. We make modifications to the networks by introducing dropout and training through MC integration based averaged loss. The comparison is made using the MNIST~\cite{mnist}, CIFAR 10~\cite{cifar10} and the CelebA datasets~\cite{celeba}. We trained the models using a batch-size of 64, iteration steps of 18750 for the MNIST dataset, 200,000 for the CIFAR-10 dataset, and 80,000 iteration steps on the CelebA. 

    \begin{figure*}
        \centering
	    \begin{subfigure}[b]{0.28\textwidth}
	        \includegraphics[width=\linewidth]{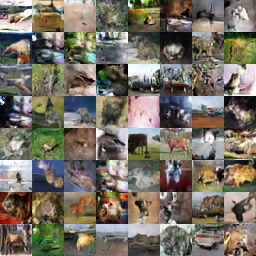}
	        \caption{NS-GAN}
	        \label{fig:vanilla_cifar10}
	    \end{subfigure}
	    \hspace{0.5cm}
	    \begin{subfigure}[b]{0.28\textwidth}
	        \includegraphics[width=\linewidth]{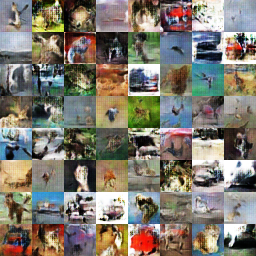}
	        \caption{LSGAN}
	        \label{fig:vanilla_cifar10}
	    \end{subfigure}
	    \hspace{0.5cm}
	    \begin{subfigure}[b]{0.28\textwidth}
	        \includegraphics[width=\linewidth]{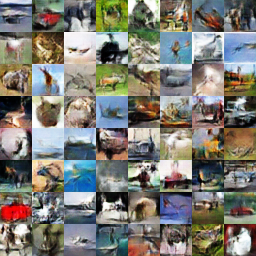}
	        \caption{WGAN}
	        \label{fig:vanilla_cifar10}
	    \end{subfigure}
	    
	    
	    \begin{subfigure}[b]{0.28\textwidth}
	        \includegraphics[width=\linewidth]{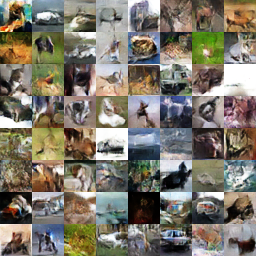}
	        \caption{Prb-NSGAN}
	        \label{fig:probabilistic_cifar10}
	    \end{subfigure}%
	    \hspace{0.5cm}
	    \begin{subfigure}[b]{0.28\textwidth}
	        \includegraphics[width=\linewidth]{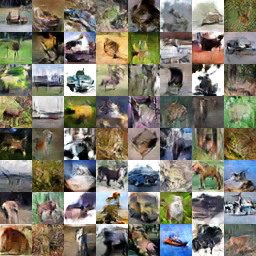}
	        \caption{Prb-LSGAN}
	        \label{fig:probabilistic_cifar10}
	    \end{subfigure}%
	    \hspace{0.5cm}
	    \begin{subfigure}[b]{0.28\textwidth}
	        \includegraphics[width=\linewidth]{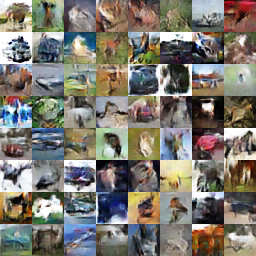}
	        \caption{Prb-WGAN}
	        \label{fig:probabilistic_cifar10}
	    \end{subfigure}%
	    \vspace{-0.5em}
	    \caption{Generated images of different variation and there probability variation GANs architectures for CIFAR 10. }\label{fig:cifar}
	    \vspace{-1.0em}
    \end{figure*}

	\begin{table}[htb]
		\centering
		\begin{tabular}{l l l l}
		\toprule
		\textbf{GAN model} & \textbf{MNIST} & \textbf{CIFAR-10} & \textbf{CelebA}\\
		\midrule
		NSGAN & 17.66 & 73.24 & 63.95 \\
		\textit{Prb}-NSGAN & 16.65 & 64.03 &\textbf{ 56.75} \\ \\
		LSGAN & 11.37 & 72.60 & 112.05 \\
		\textit{Prb}-LSGAN & \textbf{10.37} & \textbf{54.06} & 84.20 \\
		\bottomrule
		\end{tabular}
		\caption{FID scores for standard and probabilistic GANs on standard datasets. Probabilistic versions of NSGAN and LSGAN are  better with improved scores. }
		\label{tbl:probgan}
		 \vspace{-1.0em}
	\end{table}
	
	The FID score comparison for the various GAN models is given in Table~\ref{tbl:probgan}. Lower the FID score, the better the performance of the GAN. We noted that for the NSGAN and the LSGAN, our probabilistic approach improves on the FID score but not on the WGAN.
	We used the number of discriminators instances sampled $N$ = 20 and dropout probability $p$ = 0.4. Dropout was applied to both the generator and the discriminator side and the averaged loss was back-propagated. Qualitative results are presented in Figure~\ref{fig:cifar}.


	 Using the two modifications as described in section~\ref{sec:uncert}, we now implement our uncertainty measure equipped probabilistic GAN. In this section, we present the tabulation of the FID scores. The FID scores and the Inception scores are mentioned; however, note that the FID score is superior to the Inception score, and hence we emphasize the former's results. We note that in almost all cases, the use of weighted discriminator scores improves the FID score. Prb-GAN represents the probabilistic GAN based on the setup described in the previous chapter. Prb-GAN (v1) describes the probabilistic GAN setup that uses the idea of uncertainty estimation and weighted discriminator scoring. Prb-GAN (v2) is the probabilistic GAN that makes additional use of the variance of the set of discriminator logit scores. Table~\ref{tbl:uncer_fid} shows the FID score and Table~\ref{tbl:incep} the Inception scores. We found amongst all our experiments Prb-GANs equipped with uncertainty measures perform the best. We used the values $N$ = 10, and p = 0.4, where $N$ is the number of discriminators instances sampled, and p is the dropout probability. Note that we apply dropout to only the discriminator and not to the generator. 

	\begin{table}[htb]
		\centering
		\begin{tabular}{l l l l}
		\toprule
		\textbf{GAN model} & \textbf{MNIST} & \textbf{CIFAR-10} & \textbf{CelebA}\\
		\midrule
		NSGAN & 17.829 & 71.611 & 53.299 \\
		Prb-GAN & 14.749 & 58.273 & 46.355 \\
		Prb-GAN (v1) & 13.818 & 53.967 & \textbf{40.533} \\
		Prb-GAN (v2) & \textbf{13.723} & \textbf{51.040} & 40.561 \\
		\bottomrule
		\end{tabular}
		\caption{FID scores for uncertainty measure equipped GANs on standard datasets.}
		\label{tbl:uncer_fid}
		 \vspace{-1.0em}
	\end{table}

	\begin{table}[htb]
		\centering
		\begin{tabular}{l l l l}
		\toprule
		\textbf{GAN model} & \textbf{MNIST} & \textbf{CIFAR-10} & \textbf{CelebA}\\
		\midrule
		NSGAN & \textbf{2.283} & 5.290 & 1.935 \\
		Prb-GAN & 2.235 & 5.806 & 1.979 \\
		Prb-GAN (v1) & 2.258 & 6.064 & 1.962 \\
		Prb-GAN (v2) & 2.244 & \textbf{6.236} & \textbf{1.990} \\
		\bottomrule
		\end{tabular}
		\caption{Inception scores for uncertainty measure equipped GANs on standard datasets}
		\label{tbl:incep}
		\vspace{-1em}
	\end{table}
\subsection{Probabilistic Discriminators vs Probabilistic Generators}

In this section, we compare the performance of GANs that use only probabilistic discriminators against GANs that use probabilistic generators and discriminators. We find empirically that the use of probabilistic discriminators alone is sufficient to improve performance. In Table~\ref{tbl:prob_disc_vs_prob_gan}, Prb-GAN(v1) represents that Probabilistic GAN version that uses probabilistic generators and probabilistic discriminators and Prb-GAN (v2) which uses probabilistic discriminators and deterministic generators. 

	\begin{table}[htb]
		\centering
		\begin{tabular}{l l l l}
		\toprule
		\textbf{GAN model} & \textbf{MNIST} & \textbf{CIFAR-10} & \textbf{CelebA}\\
		\midrule
		
		Vanilla GAN & 17.66 & 73.24 & 63.95 \\
		Prb-GAN (v1) & 16.65 & 64.03 & 56.75 \\
		Prb-GAN (v2) & \textbf{14.75} & \textbf{58.27} & \textbf{46.35} \\
		\bottomrule
		\end{tabular}
		
		\caption{FID scores comparing GANs with probabilistic discriminators- deteriministic generators vs GANs with probabilistic generators- discriminators on standard datasets. Probabilistic discriminators alone provides superior performance as compared to probabilistic generators and discriminators. }
		\label{tbl:prob_disc_vs_prob_gan}
	\end{table}

\vspace{-2em}
\section{Conclusion}
In this paper, we have described the issues associated with the GAN model and described a probabilistic approach to solving these problems. In the first part, we discussed how the standard
technique of \textit{Dropout} could be used to create a probabilistic version of GANs and how the simple inference mechanism requiring minimal changes to current training schemes could significantly
alleviate the mode dropping issue along with training instability issues. In later sections, we described how ideas similar to the use of uncertainty measures currently used in Bayesian Neural Networks, could be
leveraged in order to improve GAN performance metrics on standard datasets further.

{\small
\bibliographystyle{IEEEtran}
\bibliography{IEEEabrv}
}

\end{document}